\documentclass[a4paper,fleqn]{cas-dc}

\usepackage[authoryear,longnamesfirst]{natbib}
\usepackage{graphicx}
\usepackage{enumitem}
 \usepackage{caption}
\usepackage{float}
\usepackage{placeins}
\usepackage{courier}
\usepackage{graphicx}
\usepackage{array}
\usepackage{booktabs, tabularx}
\usepackage{amsmath}
\usepackage{caption}
\usepackage{multirow}
\usepackage{capt-of}
\usepackage{diagbox}
\usepackage{rotating}
\usepackage{amsfonts}
\usepackage{caption}

\usepackage{hyperref}

\usepackage{titlesec}
\usepackage{enumitem}
\setlist[itemize]{leftmargin=*}
\setlist[enumerate]{leftmargin=*}

\usepackage[skins]{tcolorbox}

\newenvironment{RQ}{\vspace{2mm}\begin{tcolorbox}[enhanced,width=\linewidth,size=fbox,colback=blue!5,drop shadow southeast,sharp corners]}{\end{tcolorbox}}
\usepackage{diagbox}

\titleformat{\section}
  {\normalfont\bfseries}
  {\thesection}
  {1em}
  {}

\titlespacing*{\section}
  {0pt}
  {2.5ex}   
  {1.2ex}   

\titleformat{\subsection}
  {\normalfont\itshape}
  {\thesubsection}
  {1em}
  {}

\titlespacing*{\subsection}
  {0pt}
  {2.0ex}
  {0.9ex}

\titleformat{\subsubsection}
  {\normalfont\itshape}
  {\thesubsubsection}
  {1em}
  {}

\titlespacing*{\subsubsection}
  {0pt}
  {1.5ex}
  {0.7ex}

\begin{document}

\title[mode=title]{Modeling Image-Caption Rating from Comparative Judgments}


\author[1]{Kezia Minni}
\ead{ka4104@rit.edu}
\author[1]{Qiang Zhang}
\ead{qz8560@rit.edu}
\author[1]{Monoshiz Mahbub Khan}
\ead{mk7989@rit.edu}
\author[1]{Zhe Yu}
\ead{zxyvse@rit.edu}

\shorttitle{Comparative Image-Caption Rating}
\shortauthors{K. Minni et al.}
\affiliation[1]{
organization={Department of Software Engineering, Rochester Institute of Technology},
city={Rochester},
state={NY},
country={USA}
}

\begin{abstract}
Image caption rating is becoming increasingly important because computer-generated captions are used extensively for descriptive annotation. However, rating the accuracy of captions in describing images is time-consuming and subjective in nature. In contrast, it is often easier for people to compare (between two pairs) which image-caption pair better matches each other. In this study, we propose a machine learning framework that models such comparative judgments instead of direct ratings. The model can then be applied to rank unseen image-caption pairs in the same way as a regression model trained on direct ratings. Inspired by a state-of-the-art regression approach, we extracted visual and text features using a pre-trained ViLBERT model and tweaked the learning parameters of the baseline model to improve the model performance. This new regression model (with Kendall's $\tau_c=0.812$) outperformed the baseline model (with Kendall's $\tau_c=0.758$) on the VICR dataset. The same model structure was applied to the comparative learning framework. Trained on comparative judgments (image-caption pair A better matches each other than image-caption pair B), the comparative learning model achieved a performance similar (with Kendall's $\tau_c=0.804$) to that of the regression model. In addition, a small-scale human subject study was conducted to compare the cost and quality of direct ratings, pairwise comparisons, and same-image comparisons. The results showed that comparative judgments yielded faster results and greater agreement among human annotators than direct ratings. These results suggest that collecting comparative judgments instead of direct ratings as training data labels is promising for lower annotation costs and greater consistency. The model trained on such comparative judgments can perform as well as the model trained on direct ratings.
\end{abstract}



\maketitle

\section{Introduction}
Humans often engage in subjective decision-making, such as evaluating whether a caption accurately describes an image or whether a piece of artwork is good. Such assessments are inherently subjective and highly dependent on individual interpretations ~\cite{amabile1982social}. Automating such judgments can reduce the time and effort required for large-scale evaluations. Traditional numeric rating methods (e.g., assigning scores from 1 to 5) are often inefficient, time-consuming, and not ideal because individuals may interpret scales differently or inconsistently apply ratings, leading to noisy labels and poor inter-rater reliability~\cite{yannakakis2015ratings,zhang2020learning}.

To address these limitations, researchers have examined comparative judgment, which involves raters choosing between two options relative to each other rather than assigning scores independently. This method, based on Thurstone's Law of Comparative Judgment~\cite{thurstone1927law,roeckelein2006thurstone}, is well-suited for subjective attributes such as originality or relevance, where there is no objective ground truth available. Comparative judgment reduces individual bias and simplifies decision-making by focusing on the relative quality.

In this study, we applied comparative learning to model image-caption ratings; that is, we assessed how well a caption matched a given image. While regression models are trained on direct human ratings, which are expensive and noisy, comparative learning models are trained on comparative judgments, such as image-caption pair A better matches each other than image-caption pair B. In addition to the direct assessments, we investigated whether the comparative learning model works well in a same-image-different-caption scenario by predicting which of the two captions better matches a given image. Finally, we conducted human subject experiments to validate the advantages of comparative judgments over direct ratings. Throughout these investigations, the following research questions were explored in this study:
\begin{enumerate}[label=\textbf{ RQ\arabic*:}, leftmargin=*, nosep]
    \item \textbf{Can a model trained on comparative judgments perform as well as a regression model trained on direct ratings in predicting how well a caption matches an image?} 
    \item \textbf{Can we correctly model which caption better matches a given image?}
    \item \textbf{Can we validate the advantages of comparative judgments over direct ratings?} 
\end{enumerate}
To answer these research questions, we used multimodal modeling to compare a state-of-the-art regression model with a comparative learning model. While the regression model learns from direct human ratings to predict a single relevant value, the comparative learning model is trained to capture which image-caption pair better matches each other. Because the comparative model requires only binary decisions rather than numerical values, the training data labels can be collected more efficiently with less variation among human annotators~\cite{thurstone1927law,roeckelein2006thurstone}. 

Our findings show that comparative learning is an effective substitute for direct ratings when assessing and evaluating image–caption pairs. In RQ1, the comparative learning model performed only slightly worse ($1\%$) than the regression model with the same amount of training data  ($N$ direct ratings for the regression model vs. $N$ comparative judgments for the comparative learning model). For RQ2, comparative learning trained only on caption pairs of the same image performed well in predicting which caption better matched a given image, even when the differences between captions were subtle. In RQ3, a small-scale human subject experiment was conducted with eight human raters annotating both direct ratings and comparative judgment. The results of the experiment suggest that for the image-caption rating task, comparative judgments have shorter annotation times and higher inter-rater agreement on average than direct ratings. This result is consistent with the Law of Comparative Judgment~\cite{thurstone1927law} and other studies on comparative judgments~\cite{bramley2015investigating,verhavert2019meta}.

Our contributions are summarized as follows:
\begin{itemize}[leftmargin=*, nosep]
    \item We improved a baseline multimodal fusion regression model to accurately predict the image-caption ratings. It achieved Pearson correlation coefficient $\rho= 0.908$, Spearman's rank correlation coefficient $r_s=0.888$, and Kendall's $\tau_c=0.812$ by training on the ground-truth human ratings.
    \item To the best of our knowledge, comparative learning has not been systematically applied to solve the image-caption rating problem. It learns from comparative judgments -- image-caption pair A better matches each other than image-caption pair B or vice versa -- and predicts how well a caption describes an image. We showed that the comparative learning model achieved Pearson’s correlation coefficient $\rho= 0.876$, Spearman's rank correlation coefficient $r_s=0.883$, and Kendall's $\tau_c=0.804$ without training on the ground-truth direct ratings. This opens up the opportunity for comparative judgments, which are easier to acquire and are often more reliable than direct ratings, to be collected as training data labels instead of direct ratings.
    \item It is also shown that, comparative learning can be applied to predict which of the two captions better describes a given image with a high accuracy of $0.85$. 
    \item A small-scale human subject experiment confirmed the advantage of comparative judgments -- on average, they required shorter annotation time and achieved higher inter-rater reliability than direct ratings.  
    \item
    The code, sample dataset, and results are publicly available on GitHub\footnote{\href{https://github.com/hil-se/comparative_image_caption}{github.com/hil-se/comparative\_image\_caption}} for replication and reproduction.
\end{itemize}
\section{Background and Related Work}
\subsection{Image-Text Matching, Embeddings, and Evaluation}
Image-text matching is a multimodal training task~\cite{baltrusaitis2019multimodal,zhang2019deep}. Traditional methods usually rely on manual ratings and use regression models to directly predict matching quality. The advantages of these methods are that they are intuitive and easy to train, but they also have the disadvantage that they are "often susceptible to subjectivity and inconsistent annotations"~\cite{yannakakis2015ratings}.

A representative example is \cite{narins2024validated}, which first applied the ViLBERT pre-trained model to decompose the image and text into vectors and then used a dense neural network with two hidden layers to predict the average manual ratings. Pre-training methods such as CLIP~\cite{radford2021learning} and UNITER~\cite{chen2019uniter} also use contrastive or multi-task learning to address image-text alignment problems. However, these studies only discuss the implementation of contrastive methods and do not analyze the differences between regression and contrastive learning models.
To support such matching tasks, dual-encoder architectures have been widely adopted to learn joint image-text representations. 

\cite{radford2021learningtransferablevisualmodels} presented a dual encoder model that trains image and text representations that are compatible via contrastive learning on massive web data. The model adopts separate encoders for images and captions, thereby enabling strong zero-shot performance. We adopted a similar dual-encoder structure using ResNet for images and MiniLM for text and applied it to comparative caption evaluation tasks. 

\cite{faghri2018vseimprovingvisualsemanticembeddings} improves visual-semantic embedding models by introducing hard negative mining in a dual encoder setup for image-text retrieval. Instead of averaging over randomly sampled negatives, it selects the hardest mismatched image or caption during training, encouraging the model to make more precise distinctions. Similar to our comparative model, VSE++ uses a hinge loss objective and a dual encoder architecture, supporting our approach to learning relative caption quality. 

\cite{reimers2019sentencebertsentenceembeddingsusing} modifies BERT into a Siamese network to generate semantically meaningful sentence embeddings that can be compared using cosine similarity. MiniLM from the SentenceTransformers framework was employed by our model to encode the captions, utilizing the speed and semantic accuracy of MiniLM for comparative evaluation of the captions.

Several studies have also proposed learning-based evaluation frameworks specifically designed to assess image caption quality. TIGEr~\cite{jiang2019tiger} introduced a grounding-based framework that evaluates caption quality by measuring how well the textual content aligns with image regions. Using text-to-image grounding signals, TIGEr aims to improve the correlation with human judgment compared to text-only metrics. InfoMetIC~\cite{hu2023infometic} offers a reference-free metric for evaluating the informativeness and relevance of captions using contrastive learning. Instead of relying on ground-truth references, it estimates how well a caption captures the image content directly. QACE~\cite{lee2021qace} proposed a question-answering (QA)-based approach by generating questions about the caption and verifying consistency with the reference caption or image. Recently, CAPTURE~\cite{dong2024benchmarking} introduced a benchmark and evaluation framework designed to assess fine-grained and detailed caption quality, highlighting the limitations of existing automatic metrics.

In summary, although regression and contrastive models provide reliable baselines, they often lack robustness in capturing the subtleties of human judgment, and existing evaluation metrics are designed to correlate with human judgment rather than to learn directly from it. Our work aims to address these issues by leveraging comparative learning to better model the relative correlations and enhance generalization capabilities, directly training models to predict scalar human rating scores, and comparing regression-based learning with comparative preference learning.

\subsection{Learning from Comparative Judgments}

Learning from comparative judgments provides an alternative to direct-rating methods. Instead of directly providing scores, users compare pairs of items, which helps to obtain more consistent and detailed feedback. Comparative judgments have been shown to be effective in recommendation, information retrieval, and subjective evaluation tasks~\cite{dras2015evaluating,furnkranz2003pairwise}.

Early ranking learning models, such as RankNet and LambdaRank~\cite{burges2005learning}, use neural networks and pairwise loss functions (such as hinge or cross-entropy loss) to learn from pairwise preferences/comparative judgments. These methods have shown excellent performance in retrieval applications but have not been applied to multimodal problems.
Fürnkranz and Hüllermeier~\cite{furnkranz2003pairwise} formalized the pairwise preference learning framework by modeling ranking as a binary classification problem for a series of item pairs.
Building on this, \cite{ghosh2023adaptive} proposed an adaptive pairwise comparison method based on active sampling. Their method jointly optimizes the regression and ranking objectives, reducing annotation costs while improving preference modeling. However, their work is limited to the pure text domain and does not explore the image-text space.

\cite{khan2025efficientstorypointestimation} propose a comparative learning approach to agile story point estimation. This predicts the story point of a backlog item by training a model on a set of comparative judgments:  given two items, which one has a higher story point. A pre-trained SentenceTransformer model is first applied to embed the title and description of each item before a fully connected feedforward neural network is trained with hinge loss. The comparative learning model showed superior performance to its regression version trained on the ground truth story points. Utilizing the same comparative learning framework, \cite{bethi2026modelingartevaluationscomparative} showed that it is also beneficial to train comparative judgments in the prediction of the beauty and likings of paintings. Their human subject experiments also confirmed a $60\%$ reduction in the average annotation time when making comparative judgments instead of direct ratings.

In contrast to previous studies, this study extends the comparative learning framework to the multimodal domain. We propose a dual-encoder structure comprising a pretrained image and text embedding model, multiple dense layers, rectified linear unit (ReLU) activation functions, and dropout regularization. This proposed structure was trained using a hinge loss and demonstrated its superior ability to model human preferences and to handle subjective variability. We also compared the traditional regression model with the new comparative learning model.

\section{Methodology}\label{sec:method}
This section discusses the end-to-end approach used to train a model that predicts human preferences for image-caption pairs using both regression and comparative learning strategies. The pipeline includes data preprocessing, multimodal embedding generation, model training (regression and comparative), and evaluation.

\subsection{Notations}
Let $i$ index an image-caption pair. The multimodal representation of the $i$-th image-caption pair is denoted by $x_i$, where $x_i = [ie_i , ce_i]$ is the concatenation of the image embedding $i_e$ and the caption embedding $c_e$.

Human-provided ratings are denoted by $y_i$. A scoring function $f(\cdot)$ maps a multimodal embedding $x_i$ to a scalar utility score $f(x_i) \in \mathbb{R}$. In the regression setting, the model predicts the human ratings $\hat{y}_i = f(x_i)$.

In the comparative setting, the relative preference between two image-caption pairs $i$ and $j$ is modeled using the score difference
\[
C_{ij} = f(x_i) - f(x_j).
\]
The corresponding human preference label is denoted by $$O_{ij} =  \left\{
\begin{aligned}
&1, &\text{when } &x_i \text{ better matches each other than } x_j\\
&-1, &\text{when } &x_j \text{ better matches each other than } x_i
\end{aligned}\right..$$

For evaluation, we use Pearson ($\rho$), Spearman's rank ($r
_s$), and Kendall rank ($\tau_c$) correlation coefficients to measure the agreement between the predicted and average human-annotated scores. 


\subsection{Multimodal Embedding}
Each image-caption pair is encoded into a joint feature representation as follows:

\begin{itemize}
  \item \textbf{Visual Embedding:} A pretrained \textit{VilBERT} model is used to  process Image\_i and generate an embedding $ie_i$.
  \item \textbf{Text Embedding:} The same \textit{VilBERT} model is applied to process Caption\_i and generate an embedding $ce_i$.
\end{itemize}

The image and text embeddings are then concatenated to form a unified multimodal vector for each image-caption pair $x_i = [ie_i , ce_i]$.


\begin{figure*}[!tbh]
    \centering

    \includegraphics[width=\textwidth]{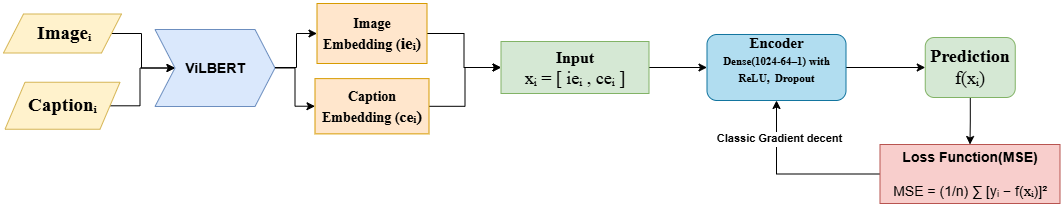}
    \captionof{figure}{Regression framework showing preprocessing, multimodal fusion, and model training path~\cite{narins2024validated}.}
    \label{fig:regression_model}

    \includegraphics[width=\textwidth]{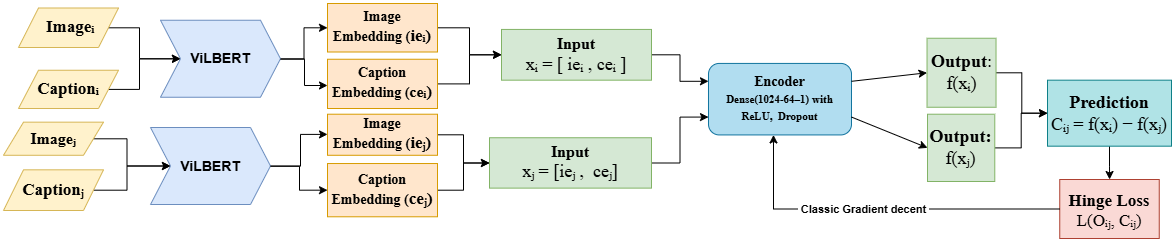}
    \captionof{figure}{Comparative learning framework with hinge loss to learn preferences between image-caption pairs.}
    \label{fig:pairwise_model}
\end{figure*}

\subsection{Baseline Regression Modeling}

First, we adopted a state-of-the-art regression architecture~\cite{narins2024validated} that predicts the quality of the caption as a continuous scalar value. As shown in 
\hyperref[fig:regression_model]{Fig.~\ref*{fig:regression_model}}, this regression model utilizes the pretrained VilBERT model to generate the image and caption embeddings and concatenate them as the input to an encoder with three fully connected dense layers. Each hidden layer used a rectified linear unit (ReLU) activation and dropout. The regression model was trained using the Adaptive Moment Estimation (Adam) optimizer to minimize the mean squared error (MSE). In contrast to \cite{narins2024validated}, which used a dropout rate of $80\%$ and a learning rate of $10^{-5}$, decayed by $1\%$ every 15 epochs, we used a dropout rate of $20\%$ and a learning rate of $10^{-3}$ with cosine decay of \texttt{decay\_steps}$=5000$ 
and \texttt{alpha}$=0.0001$. To prevent overfitting, we also applied a \texttt{ReduceLROnPlateau} 
callback (monitoring validation loss with \texttt{patience} $= 8$, \texttt{factor} $= 0.3$, 
and \texttt{min\_lr} $= 10^{-6}$) and an \texttt{EarlyStopping} callback (monitoring 
validation loss with \texttt{patience} $= 15$, restoring the best weights upon termination).

\subsection{Comparative Learning Modeling}

We then applied the same encoder to form a comparative learning model that learns to estimate caption quality based on comparative judgments between two image and caption pairs. As shown in \hyperref[fig:pairwise_model]{Fig.~\ref*{fig:pairwise_model}}, the model takes a pair of inputs $(x_i,x_j)$ every time and compares their outputs $f(x_i)$ and $f(x_j)$ from the encoder:
\[
C_{ij} = f(x_i) - f(x_j)
\]
An Adam optimizer was then applied to minimize the hinge loss:
\[
\mathcal{L}(O_{ij}, C_{ij}) = \max(0, 1 - O_{ij} \cdot C_{ij})
\]
Where $O_{ij}$ denotes the comparative preference label between the two image-caption pair inputs $(x_i, x_j)$:
\begin{equation*}
O_{ij} =  \left\{
\begin{aligned}
&1, &\text{when } &x_i \text{ better matches each other than } x_j\\
&-1, &\text{when } &x_j \text{ better matches each other than } x_i.
\end{aligned}\right.
\end{equation*}

This comparative learning model closely mimics the manner in which humans make subjective decisions in areas such as caption evaluation. It is often easier for a rater to judge which caption better suits the image than to assign an exact numeric score. Therefore, comparative learning serves as a supervision signal more in line with human nature for more in line with subjective quality estimation. Because the model is trained to learn relative preferences, it avoids issues related to scale inconsistency, which are often observed in direct ratings. This makes the collection of comparative judgments much faster and less noisy, while still conveying fine-grain preference information. 


Overall, the comparative model provides an interpretable utility-based framework that learns human-like preference ordering and generalizes well to new image-caption pairs.

\section{Experiments}
\subsection{Data}

We used the Validated Image Caption Rating (VICR) dataset \cite{narins2024validated}, which consists of 15,646 image-caption pairs annotated with 68,217 human ratings. Each image-caption pair was rated on a scale of 1 to 5, with 1 being the least liked.

The dataset includes a mix of human-written and machine-generated captions, providing a wide range of quality levels suitable for both regression and comparative learning models. Each image-caption pair received ratings from multiple annotators, and we computed the mean rating per pair as the ground truth for regression and to derive relative preference labels for comparative learning. \hyperref[tab:data_example]{Table~\ref*{tab:data_example}} shows four example image-caption pairs from the dataset.

\begin{table}[!tbh]
\caption{Example of image-caption pairs with their average human ratings $y$, which denotes the ground truth rating from the dataset.}
\label{tab:data_example}
\centering
\renewcommand{\arraystretch}{1.25}
\begin{tabular}{ |>{\centering\arraybackslash}m{3 cm} |
                 >{\centering\arraybackslash}m{2.8 cm}| 
                >{\centering\arraybackslash}m{1.0 cm}| }
\hline
\textbf{Image} & \textbf{Caption} & \textbf{$y_i$} \\
\hline
\includegraphics[width=\linewidth]{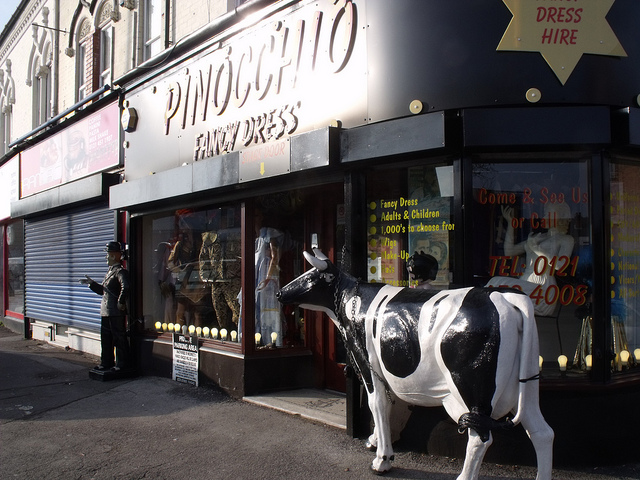} 
& A woman in a red jacket standing next to a sign. 
& 1.6 \\
\hline
\includegraphics[width=\linewidth]{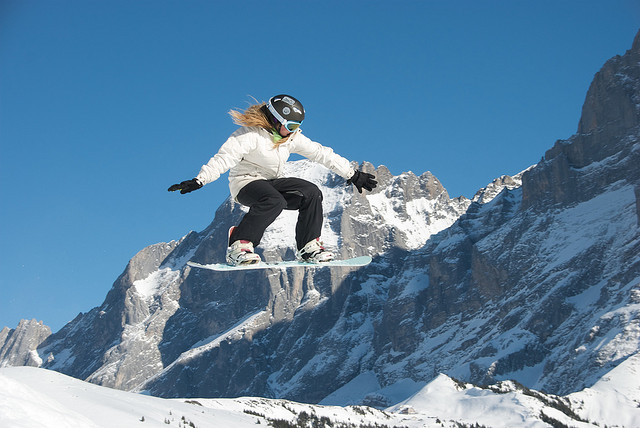} 
& A man in a black jacket is walking down a hill. 
& 2.0 \\
\hline
\includegraphics[width=\linewidth]{IMG8.jpg} 
& People walking down the street in the rain. 
& 2.75 \\
\hline

\includegraphics[width=\linewidth]{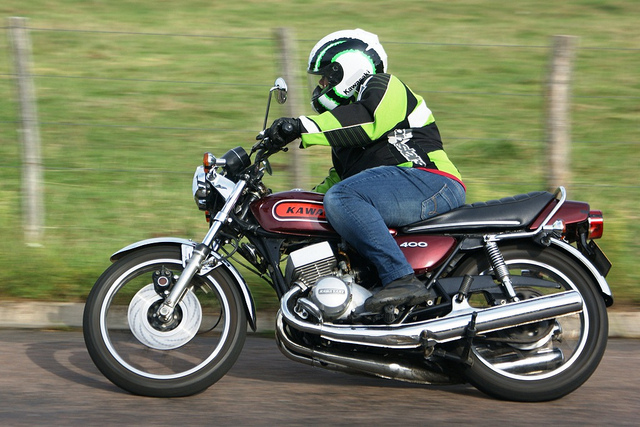} 
& A man on a motorcycle going down the street.
& 5.0 \\

\hline
\end{tabular}
\end{table}

\begin{table}[ht]
\caption{\small Example of comparative judgments between two image-caption pairs with average ratings $y_i$ and $y_j$ (e.g., $y_i = 1.5$, $y_j = 3.75$), where $y$ denotes the average ground-truth rating from the dataset, resulting in a comparative label $O_{ij} = \mathrm{sgn}(y_i-y_j)$ used for hinge loss training.}
\label{tab:comparison_label_example}
\centering
\renewcommand{\arraystretch}{1.25}
\begin{tabular}{| >{\centering\arraybackslash}m{2.0cm}| 
                >{\centering\arraybackslash}m{2.5cm} |
                 >{\centering\arraybackslash}m{1.0cm} |
                >{\centering\arraybackslash}m{1.0cm} |}
\hline
\textbf{Image} & \textbf{Caption} & \textbf{\centering $y_i$} & \textbf{\centering $O_{ij}$} \\
\hline
\includegraphics[width=2.0cm,height=4.0cm,keepaspectratio]{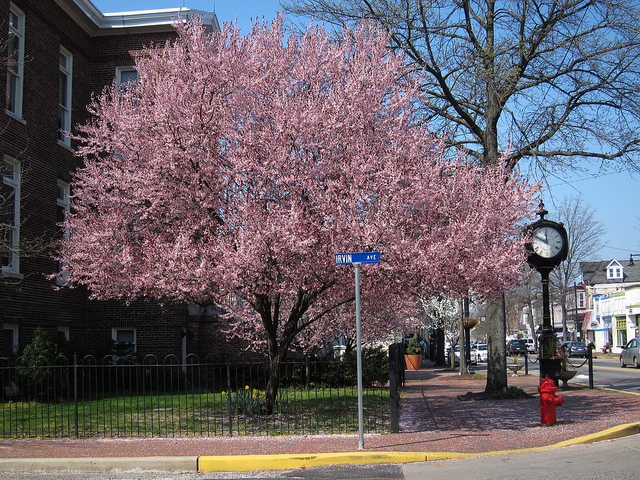}
& A table and chairs are placed on a grassy walkway with a pitcher. 
& 1.5 
& \multirow{2}{*}{\centering \textbf{-1}} \\
\cline{1-3}
\includegraphics[width=2.0cm,height=4.0cm,keepaspectratio]{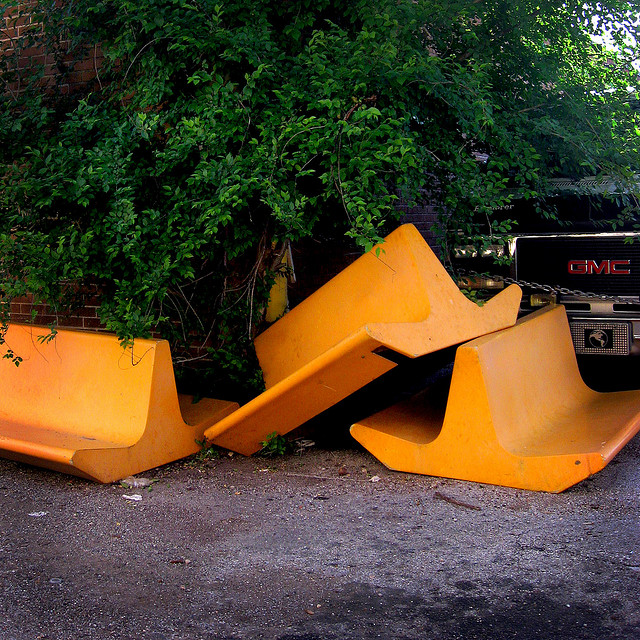}
&Orange construction equipment sitting on the side of the road 
& 3.75 
& \\
\hline
\includegraphics[width=2.0cm,height=4.0cm,keepaspectratio]{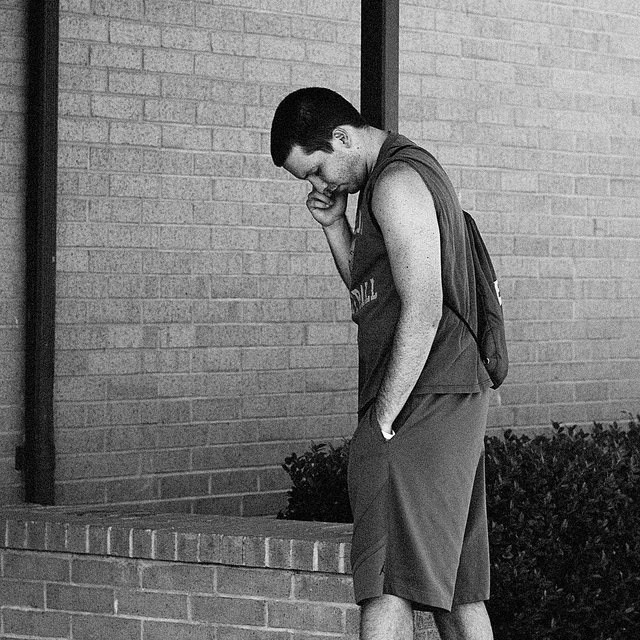} 
& A man standing in front of a brick wall. 
& 4.25 
& \multirow{2}{*}{\centering \textbf{+1}} \\
\cline{1-3}
\includegraphics[width=2.0cm,height=4.0cm,keepaspectratio]{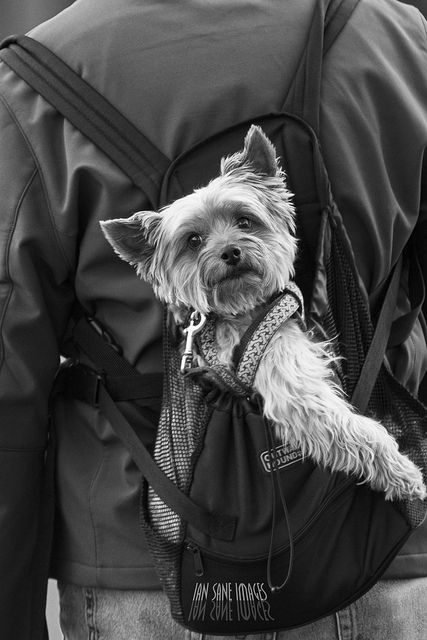} 
& A laptop next to two computers, keyboards and some mice. 
& 1.0 
& \\
\hline
\end{tabular}
\end{table}

\hyperref[tab:comparison_label_example]{Table~\ref*{tab:comparison_label_example}}
shows how comparative training examples are generated from the VICR dataset using image-caption pairs associated with different images. For each pair $(x_i, x_j)$, the normalized human ratings $y_i$ and $y_j$ are compared to derive a binary preference label $O_{ij} = \mathrm{sgn}(y_i-y_j)$. 

\begin{table}[ht]
\caption{\small Same-image, different-caption pairs from the VICR dataset with average ratings $y_i$ and $y_j$, where $y$ denotes the ground-truth rating, resulting in a comparative judgment $O_{ij} = \mathrm{sgn}(y_i - y_j)$ used for hinge loss training.}
\label{tab:same_image_pairs}
\centering
\renewcommand{\arraystretch}{2}
\setlength{\extrarowheight}{2pt}
\setlength{\tabcolsep}{4pt}
\resizebox{1.0\linewidth}{!}{
\begin{tabular}{| >{\centering\arraybackslash}m{2cm} 
                 |>{\centering\arraybackslash}m{2.5cm} 
                 |>{\centering\arraybackslash}m{1.0cm} 
                 |>{\centering\arraybackslash}m{1 cm} |}
\hline
\small
\textbf{Image} & \textbf{Caption} & \textbf{\centering $y_i$} & \textbf{\centering $O^{\text{same}}_{ij}$} \\
\hline
\multirow{2}{*}{\includegraphics[width=\linewidth]{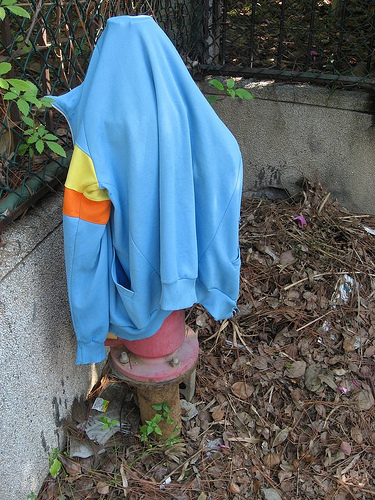}} 
& A blue umbrella sitting in the ground next to a fence.
 & 2.8 & \multirow{2}{*}{\centering \textbf{-1}} \\
\cline{2-3}
& A fire hydrant with a blue jacket on top of it. & 4.2 & \\
\hline
\multirow{2}{*}{\includegraphics[width=\linewidth]{SS3.jpg}} 
& Mom gives her daughter a lesson in using her baseball glove. & 4.75 & \multirow{2}{*}{\centering \textbf{+1}} \\
\cline{2-3}
& A group of young people playing a game of baseball. & 3.25 & \\
\hline
\end{tabular}
}

\end{table}

\hyperref[tab:same_image_pairs]{Table~\ref*{tab:same_image_pairs}} shows a subset of the dataset where each image has multiple captions. With this subset of data, we explore RQ2 by finding the best matching caption for a given image by only generating comparative judgments between the captions of the same image.

\subsection{RQ1: Regression VS. Comparative Learning}\label{sec:RQ1}

For RQ1, we compared the performance of image-caption rating predictions of the proposed models from Section~\ref{sec:method} on the Validated Image Caption Rating (VICR) dataset~\cite{narins2024validated}. Same for all treatments, the last 20\% of the data $X^\text{test}$ were held out for testing, 64\% were used for training $X^\text{train}$, and 16\% for validation $X^\text{val}$. 
\begin{itemize}
\item
\cite{narins2024validated}: The same result from \cite{narins2024validated} was used as our baseline. 
\item
Regression: As Shown in \hyperref[fig:regression_model]{Fig.~\ref*{fig:regression_model}}, we applied the same architecture as \cite{narins2024validated} but with different hyperparameters. 
\item
Comparative: As Shown in \hyperref[fig:pairwise_model]{Fig.~\ref*{fig:pairwise_model}}, we incorporated the regression model into a comparative learning framework to train comparative judgments instead of direct ratings. These comparative judgments were generated using the training and validation data $X^\text{train}_{ij} = \{(x_i, x_j)|x_i,x_j\in X^\text{train}\}$, $X^\text{val}_{ij} = \{(x_i, x_j)|x_i,x_j\in X^\text{val}\}$. The number of data points used was the same as in the regression model: $|X^\text{train}_{ij}|=|X^\text{train}|=9,988$, $|X^\text{val}_{ij}|=|X^\text{val}|=2,497$.
\end{itemize}

\subsubsection{Training and Evaluation for RQ1: }

All models were implemented in TensorFlow and trained using a GeForce RTX 4090 GPU. The performances of how well the predicted ratings $f(x)$ correlate to the ground truth ratings $y$ were evaluated using
\begin{itemize}
  \item Pearson correlation $\rho$
  \begin{equation}
    \rho = \frac{\sum_{i=1}^{n} (y_i - \bar{y})(f(x_i) - \overline{f(x)})}{\sqrt{\sum_{i=1}^{n} (y_i - \bar{y})^2} \sqrt{\sum_{i=1}^{n} (f(x_i) - \overline{f(x)})^2}}
\end{equation}
  \item Spearman rank correlation $r_s$
  \begin{equation}
    { r_{s}= {\rho } {\bigl [}R[y], R[f(x)]\ {\bigr ]}},
\end{equation}
where $R[y]$ and $R[f(x)]$ are the ranks of $y$ and $f(x)$.
\item Kendall's $\tau_c$
  \begin{equation}
    \tau _{c}={\frac {2(n_{c}-n_{d})}{n^{2}{\frac {(m-1)}{m}}}},
\end{equation}
where
\begin{equation*}
\begin{aligned}
    n_{c} = &\text{Number of concordant pairs: }\\
    &|\{(i,j)|\text{sgn}(y_i-y_j) = \text{sgn}(f(x_i)-f(x_j))\}|,\\
    n_{d} = &\text{Number of discordant  pairs: }\\
    &|\{(i,j)|\text{sgn}(y_i-y_j) \neq \text{sgn}(f(x_i)-f(x_j))\}|,\\
    r = &\text{Number of distinct }y_i,\\
    c = &\text{Number of distinct }f(x_i),\\
    m = &\min(r, c).
\end{aligned}
\end{equation*}
\end{itemize}
Note that \cite{narins2024validated} only reported Kendall's $\tau_c$.

To statistically measure the difference, we repeated each experiment five times and applied Student's t test~\cite{owen1965power} to measure whether the differences in the metrics were significant. We also applied Cohen's d~\cite{diener2010cohen} to test the effect sizes of the differences.



\subsection{RQ2: Which caption better matches a given image?}

In addition to the ratings of a given image-caption pair, it is also a common task to determine which caption better describes a given image. RQ2 focuses on this aspect. For this experiment, we used a subset of the data in which one image was associated with multiple captions from the original training, validation, and test sets. Then, comparative judgments were generated as $X^\text{same}_{ij} = \{(x_i, x_j)|\text{image}_i = \text{image}_j\}$. Therefore, we have $X^\text{s\_train}_{ij} = \{(x_i, x_j)|\text{image}_i = \text{image}_j,\,\text{image}_i\in X^\text{train}\}$, $X^\text{s\_val}_{ij} = \{(x_i, x_j)|\text{image}_i = \text{image}_j,\,\text{image}_i\in X^\text{val}\}$, and $X^\text{s\_test}_{ij} = \{(x_i, x_j)|\text{image}_i = \text{image}_j,\,\text{image}_i\in X^\text{test}\}$. Each set has $|X^\text{s\_train}_{ij}|=5,066$, $|X^\text{s\_val}_{ij}|=1,073$, and $|X^\text{s\_test}_{ij}|=1,701$ image-caption pairs, respectively.

Three different treatments were evaluated for RQ2.
\begin{itemize}
\item
Regression: The same regression model from RQ1 trained on $X^\text{train}$ and validated on $X^\text{val}$.
\item
Comparative: the same comparative learning model from RQ1 trained on $X_{ij}^\text{train}$ and validated on $X_{ij}^\text{val}$.
\item
Same Image: the comparative learning model trained and validated only on comparative judgments between captions describing the same image $X^\text{s\_train}_{ij}$ and $X^\text{s\_val}_{ij}$.
\end{itemize}

The performance of each treatment was evaluated on the test set $X^\text{s\_test}_{ij}$ with accuracy:
  \begin{equation}
    \text{Accuracy}=\frac{|\{(x_i, x_j)|C_{ij}=O_{ij},\,(x_i, x_j)\in X^\text{s\_test}_{ij}\}|}{|X^\text{s\_test}_{ij}|}.
\end{equation}

\subsection{RQ3: Human Evaluation Efficiency Study}

RQ3 explored whether the advantages of comparative judgments over direct ratings in image-caption rating could be confirmed with small-scale human subject experiments. We conducted a controlled annotation study in which eight participants completed three tasks designed to mirror real-world subjective caption evaluation scenarios after obtaining informed consent.
\begin{enumerate}
    \item \textbf{Task 1: }Assign quality scores to individual image–caption pairs as shown in \hyperref[fig:task1]{Fig.~\ref*{fig:task1}},
    \item \textbf{Task 2: }Choose which image-caption pair better matches each other as shown in \hyperref[fig:task2]{Fig.~\ref*{fig:task2}},
    \item \textbf{Task 3: } Choose which caption better matches a given image as shown in \hyperref[fig:task3]{Fig.~\ref*{fig:task3}}.
\end{enumerate}



\begin{center}
\includegraphics[width=0.85\linewidth]{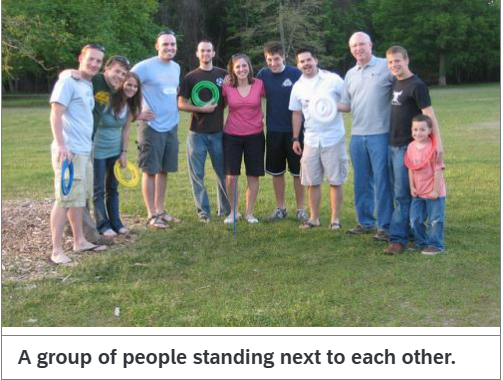}
\captionof{figure}{Task 1: Direct rating of each caption on a 1–5 scale. Participants judged how well the caption described the corresponding image.}
\label{fig:task1}
\end{center}

\begin{center}
\includegraphics[width=0.85\linewidth]{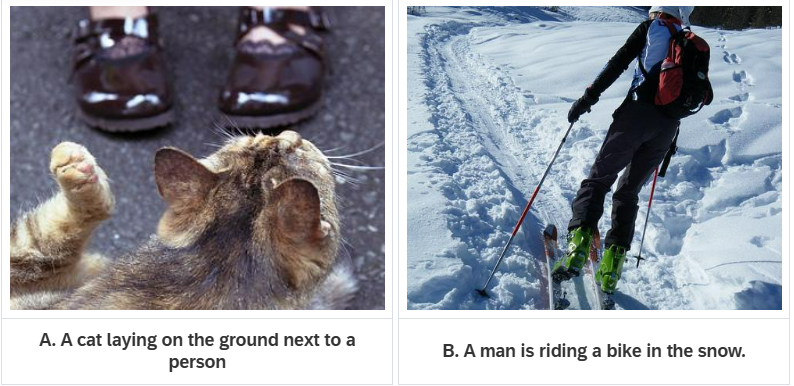}
\captionof{figure}{Task 2: Pairwise comparison across different images. Participants selected which image-caption pair better matches each other.}
\label{fig:task2}
\end{center}

\begin{center}
\includegraphics[width=0.85\linewidth]{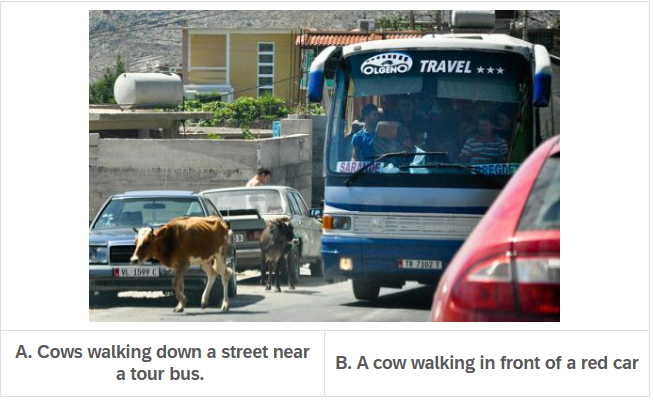}
\captionof{figure}{Task 3: Pairwise comparison of captions describing the \textit{same image}. Participants selected which caption better matches the given image.}
\label{fig:task3}
\end{center}

For each participant and task, we recorded the time taken for each question and the option selected. This allows us to compare the annotation speed and inter-rater agreement across the three tasks: direct rating, pairwise comparison (between different image-caption pairs), and same-image comparison. While the inter-rater agreement for Task 1 can be evaluated with $\rho$ and $r_s$, the inter-rater agreements for Task 2 and 3 are evaluated using
\begin{itemize}
    \item Observed agreement $p_o$
        \begin{equation}
            p_o = \frac{1}{n}|R_{ij}^{(p)} = R_{ij}^{(q)}|
        \end{equation}
        where $R_{ij}^{(p)}$ is the comparative judgment on the item pair $(x_i, x_j)$ from rater $p$ and $n=10$ is the total number of questions.
    \item Expected agreement $p_e$
        \begin{equation}
            p_e = \frac{1}{n^2}\sum_{k=0}^1 |R_{ij}^{(p)}=k|\cdot |R_{ij}^{(q)}=k|
        \end{equation}
    \item Cohen’s $\kappa$
        \begin{equation}
            \kappa = \frac{p_o - p_e}{1 - p_e}
        \end{equation}

\end{itemize}

\begin{table}[!tbh]
\caption{Performance metrics when using 80\% for training and validation and 20\% for testing. The comparative learning model used same number of comparative judgments generated from the 80\% training data.}
\label{tbl:regression_metrics}
\begin{tabular*}{\tblwidth}{@{}LCCC@{}}
\toprule
\textbf{Treatment} & \textbf{$\rho$} & \textbf{$r_s$} & \textbf{$\tau_c$} \\
\midrule
 \cite{narins2024validated} & - & - & $0.758\pm0.03$ \\
Regression & \textbf{$0.908\pm0.001$} & \textbf{$0.887\pm0.001$} &\textbf{$0.811\pm0.001$}  \\
Comparative & \textbf{$0.874\pm0.008$} & \textbf{$0.880\pm0.002$} &\textbf{$0.800\pm0.002$}  \\
\bottomrule
\end{tabular*}
\end{table}

\section{Results}
This section presents the quantitative findings based on the three research questions addressed in this study. We first compare the regression and comparative models (RQ1), then examine the same-image caption comparison setting (RQ2), and finally report the results of the human evaluation (RQ3).

\subsection{RQ1: Regression vs. Comparative Learning}

\subsubsection{Comparison to the state-of-the-art baseline:}
\hyperref[tbl:regression_metrics]{Table~\ref*{tbl:regression_metrics}} shows the results of the three treatments discussed in Section~\ref{sec:RQ1} including the state-of-the-art baseline from \cite{narins2024validated}. To ensure a fair comparison, all three treatments used the same number of training and validation data: the two regression models were trained and validated on $|X^\text{train}|=9,988$ and $|X^\text{val}|=2,497$ image-caption pairs, while the comparative learning model was trained and validated on $|X_{ij}^\text{train}|=9,988$ and $|X_{ij}^\text{val}|=2,497$ randomly sampled data pairs from the corresponding training and validation set. From \hyperref[tbl:regression_metrics]{Table~\ref*{tbl:regression_metrics}}, we can observe the following:
\begin{itemize}
    \item Our improved version of the regression model outperformed the baseline configuration from \cite{narins2024validated} by $7\%$, improving the $\tau_c$ from $0.758$ to $0.811$. The Student's t-test also confirmed that this difference was statistically significant (rejecting the null hypothesis with $p=0.004<0.05$). Cohen's $d=0.426$ indicates a small to medium effect size.
    \item The comparative learning model also performed worse than the regression model in terms of $\rho$ by $4\%$, in terms of $r_s$ by $1\%$, and in terms of $\tau_c$ by $1\%$. The t test confirmed that the differences are significant with $p=0.000$, and the effect sizes are huge based on Cohen's $d>5$.  
\end{itemize}
Although the results in \hyperref[tbl:regression_metrics]{Table~\ref*{tbl:regression_metrics}} shows that the comparative learning model underperformed the regression model, their differences are small ($1\%$) especially for $r_s$ and $\tau_c$. This is because $r_s$ and $\tau_c$ only measure whether the order/ranks are consistent while $\rho$ measures whether there is a linear correlation. Given the nature of comparative judgments, the comparative learning model can only learn the relative order/ranks instead of the scales.

\subsubsection{How training data size affects the performance:}
Here, we also evaluated the performances of the regression and comparative learning models with different number of training and validation data. That is, the regression model was trained and validated on $k\cdot N$ randomly sampled image-caption ratings from the original training and validation set where $k\in\{0.1,0.2,0.3,0.4,0.5,0.6,0.7,0.8,0.9,1\}$ and $N=12,485$. The comparative learning model was trained and validated on $k\cdot N$ comparative judgments randomly generated from the original training and validation set where $k\in\{0.1,0.2,0.3,0.4,0.5,0.6,0.7,0.8,0.9,1, 2,3,4,5\}$ and $N=12,485$. Each experiment was repeated for 5 times and the average performances were reported in \hyperref[fig:comparison_limited]{Fig.~\ref*{fig:comparison_limited}}:
\begin{itemize}
\item
The comparative learning model achieved its best performance at $k=0.6$ with $\rho=0.877$, $r_s=0.881$, and $\tau_c=0.802$. This is close to the performance of the regression model at $k=0.6$ with $\rho=0.905$, $r_s=0.884$, and $\tau_c=0.807$.
\item
The regression model achieved its best performance at $k=0.9$ with $\rho=0.908$, $r_s=0.888$, and $\tau_c=0.811$.
\end{itemize}
These results suggest that, while the regression model benefits from more training data, increasing the number of training data pairs does not necessarily improve the performance of the comparative learning model. It also suggests that the annotation cost could be further reduced to $k\cdot N$ comparative judgments.

\begin{center}
    \includegraphics[width=0.5\textwidth]{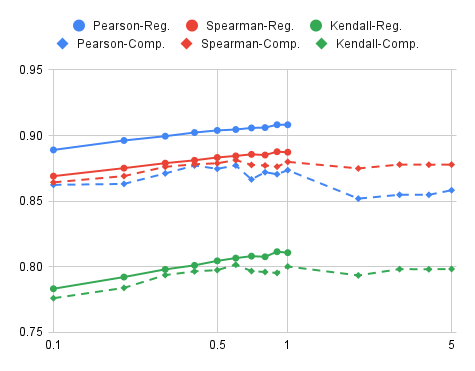}
    \captionof{figure}{Pearson $\rho$, Spearman's $r_s$, and Kendall's $\tau_c$ for the regression model and the comparative learning model with different number of training and validation data. $k=1$ means all the 80\% training and validation data is used.}
    \label{fig:comparison_limited}
\end{center}

\begin{RQ}\textbf{Answer to RQ1: }Our regression model outperformed the baseline configuration by $7\%$ in terms of $\tau_c$. The comparative learning model trained and validated on 12,485 comparative judgments performed only slightly worse $1\%$ than its regression counterpart trained and validated on 12,485 image-caption ratings. This shows promising effort reduction in training data annotation since comparative judgments are easier to acquire than direct ratings. The comparative learning model has the potential to further reduce its annotation cost by only using $12,485\times 0.6 = 7,491$ comparative judgments for training and validation.
\end{RQ}


\subsection{RQ2: Same-Image Caption Preference Modeling}

\subsubsection{Comparative Learning Performance for Same Image Different Captions:}

It is also a common task to determine which caption better describes a given image.

\begin{table}[ht]
\caption{Same-image different-caption model evaluation averaged in 5 runs. Each treatment was evaluated for its accuracy in predicting which test caption better describes a given image on the same test data. ``Regression'' represents the same regression model from \hyperref[tbl:regression_metrics]{Table~\ref*{tbl:regression_metrics}} in RQ1, ``Comparative'' represents the same comparative learning model from \hyperref[tbl:regression_metrics]{Table~\ref*{tbl:regression_metrics}} in RQ1, and ``Same Image'' represents the comparative learning model trained only on comparative judgments between captions describing the same image.}
\label{tbl:sameimage_results}
\begin{tabular*}{\tblwidth}{@{}LCCC@{}}
\toprule
\textbf{Treatment} & \textbf{Accuracy}  \\
\midrule
Regression &  $0.857\pm0.004$ \\
Comparative &  $0.846\pm0.009$ \\
\textbf{Same Image} &  $0.848\pm0.004$ \\
\bottomrule
\end{tabular*}
\end{table}

Table~\ref{tbl:sameimage_results} compares the average accuracy of three treatments in predicting which test caption better describes a given image on the same test data. While the ``Regression'' model has the highest accuracy, the ``Same Image'' model has similar accuracy with the ``Comparative'' model.

The null hypothesis of the accuracy of ``Comparative'' and ``Same Image'' being the same was accepted by the Student's t-test with a p value $=0.709>0.05$. So there is no significant difference in the performance of the ``Comparative'' model and the ``Same Image'' model. For this task, the ``Same Image'' model has an advantage of only need to train on comparative judgments between captions describing the same image.

The null hypothesis of the accuracy of ``Regression'' and ``Same Image'' being the same was rejected by the Student's t-test with a p value $=0.006<0.05$. The effect size is also large with Cohen's d $=2.336$. So the performance of the regression model is still significantly better than the ``Same Image'' model for this task. However, their performance difference is only $1\%$ based on the average accuracy while the required annotation cost could be very different.

\begin{RQ}\textbf{Answer to RQ2: }The comparative learning model trained and validated on 6,139 caption pairs describing the same image achieved similar performance as the comparative learning model trained and validated on 12,485 comparative judgments (most made between captions describing different images) in terms of predicting which caption better matches a given image. It also achieved only $1\%$ worse accuracy when compared to the regression model trained and validated on 12,485 image-caption ratings. This shows promising effort reduction in training data annotation.
\end{RQ}

\subsection{RQ3: Human Evaluation Results}

In addition to model-based evaluation, we conducted human-based experiments with eight participants performing three tasks: 1) assigning scores to individual image-caption pairs, 2) pairwise comparison between two image and caption pairs, and 3) comparison between two captions describing the same image. The detailed results of these three tasks are presented below.

\subsubsection{Task 1: Direct Rating Results}

Participants were assigned scores between 1 and 5 based on how relevant each caption was to its corresponding image. The responses were recorded in \hyperref[tbl:task1]{Table~\ref*{tbl:task1}}.

\begin{table}
\caption{Task 1 direct rating summary. Each row corresponds to a single image caption pair evaluated using direct ratings on a 1–5 scale. $y_i$ denotes the ground truth score from the VICR dataset. $R_i^{(1)}$ to $R_i^{(8)}$ represent ratings provided by eight participants. $\bar{R}_i$ denotes the mean participant rating for each item.}
\label{tbl:task1}
\begin{tabular*}{\tblwidth}{@{}LCCCCCCCCCC@{}}
\toprule
Q & $y_i$ & $R_i^{(1)}$ & $R_i^{(2)}$ & $R_i^{(3)}$ & $R_i^{(4)}$ & $R_i^{(5)}$ & $R_i^{(6)}$ & $R_i^{(7)}$ & $R_i^{(8)}$ & $\bar{R}_i$ \\
\midrule
Q1  & \textbf{2.4}  & 3 & 3 & 4 & 1 & 2 & 2 & 1 & 1 & \textbf{2.13} \\
Q2  & \textbf{1.6}  & 1 & 1 & 1 & 1 & 1 & 1 & 1 & 1 & \textbf{1.0} \\
Q3  & \textbf{2.0}  & 2 & 2 & 1 & 5 & 1 & 2 & 1 & 1 & \textbf{1.88} \\
Q4  & \textbf{5.0}  & 5 & 5 & 3 & 5 & 5 & 4 & 2 & 5 & \textbf{4.25} \\
Q5  & \textbf{3.0}  & 3 & 3 & 2 & 2 & 2 & 3 & 3 & 1 & \textbf{2.38} \\
Q6  & \textbf{2.75} & 2 & 1 & 2 & 1 & 1 & 2 & 5 & 5 & \textbf{2.38} \\
Q7  & \textbf{2.5}  & 1 & 1 & 2 & 1 & 2 & 1 & 1 & 1 & \textbf{1.25} \\
Q8  & \textbf{2.75} & 3 & 2 & 3 & 3 & 2 & 1 & 1 & 1 & \textbf{2.0} \\
Q9  & \textbf{1.0}  & 1 & 1 & 1 & 1 & 1 & 1 & 1 & 1 & \textbf{1.0} \\
Q10 & \textbf{4.5}  & 5 & 5 & 5 & 5 & 5 & 5 & 5 & 5 & \textbf{5.0} \\
\bottomrule
\end{tabular*}
\end{table}

\textit{{Task 1: Agreement between the ground truth and average ratings}}
\begin{itemize}
    \item $\text{MAE}(y_i, \bar{R}_i) = 0.525$
    \item $\rho(y_i, \bar{R}_i)= 0.931$
    \item $r_s(y_i, \bar{R}_i)= 0.908$.
\end{itemize}
Here, the average participant rating refers to the mean rating across the eight annotators in our human evaluation, reported as the $\bar{R}_i$ column in \hyperref[tbl:task1]{Table~\ref*{tbl:task1}}. The ground truth corresponds to the average ratings provided across all annotators in the original dataset and is shown as the $y_i$ column in \hyperref[tbl:task1]{Table~\ref*{tbl:task1}}.

The average participant rating achieved an MAE of 0.525 against the ground-truth score, with Pearson  $\rho = 0.931$ and Spearman $r_s = 0.908$. While these metrics indicated reasonable alignment with the true ratings, agreement among participants was noticeably lower. Inter-rater correlations (Pearson $\rho = 0.654$, Spearman $r_s = 0.613$) revealed considerable variability in how individuals applied the rating scale, even though a majority agreement was reached for all items. This suggests that direct ratings introduce subjective differences among annotators.

\textit{{Task 1: Inter-Rater Reliability}}
\begin{itemize}
    \item \textbf{Average inter-rater Pearson ($\rho$) (across raters):} $$\overline{\rho(R_i^{(p)}, R_i^{(q)})}= 0.654.$$
    \item \textbf{Average inter-rater Spearman ($r_s$) (across raters):} $$\overline{r_s(R_i^{(p)}, R_i^{(q)})}= 0.613.$$
\end{itemize}
Inter-rater reliability was assessed to evaluate the consistency with which raters used the direct rating scale. While the mean rater ratings demonstrated a high level of agreement with the ground truth, these correlation figures indicated considerable variation in how different annotators applied the scoring rubric. The moderate inter-rater Pearson and Spearman correlations indicate that the participants showed reasonable consistency in their rating behavior, although their exact numerical scores were not identical. This shows that direct rating tasks can produce inconsistent individual ratings, even when the overall average is closer to the ground truth. Inter-rater reliability was computed by measuring Pearson and Spearman correlations between all unique rater pairs, such as $R_i^{(1)}$ vs. $R_i^{(2)}$ and $R_i^{(4)}$ vs. $R_i^{(7)}$, and then averaging the results across all pairs.

\subsection{Task 2: Pairwise Comparison}

Participants were assigned to compare a given pair of two image-caption pairs for which the pair better matched each other. The responses were recorded in \hyperref[tbl:task2]{Table~\ref*{tbl:task2}}.

\begin{table}
\caption{Task 2 pairwise comparison summary. $y_i$ and $y_j$ denote the ground truth ratings for $x_i$ and $x_j$, respectively. $R_{ij}^{(1)}$ to $R_{ij}^{(8)}$ indicate individual annotator preferences, where +1 denotes a preference for $x_i$ and -1 denotes a preference for $x_j$. $M_{ij}$ represents the aggregated human preference.}
\label{tbl:task2}
\begin{tabular*}{\tblwidth}{@{}LCCCCCCCCCCC@{}}
\toprule
Q & $y_i$ & $y_j$ & $R_{ij}^{(1)}$ & $R_{ij}^{(2)}$ & $R_{ij}^{(3)}$ & $R_{ij}^{(4)}$ & $R_{ij}^{(5)}$ & $R_{ij}^{(6)}$ & $R_{ij}^{(7)}$ & $R_{ij}^{(8)}$ & $M_{ij}$ \\
\midrule
Q1  & \textbf{1.5}  & \textbf{3.75} & -1 & -1 & \cellcolor{gray!25}+1 & -1 & -1 & -1 & -1 & -1 & \textbf{-1} \\
Q2  & \textbf{4.4}  & \textbf{2.5}  & +1 & +1 & +1 & +1 & +1 & +1 & +1 & +1 & \textbf{+1} \\
Q3  & \textbf{4.25} & \textbf{1.0}  & +1 & +1 & +1 & +1 & +1 & +1 & +1 & +1 & \textbf{+1} \\
Q4  & \textbf{5.0}  & \textbf{1.0}  & +1 & +1 & +1 & +1 & +1 & +1 & +1 & +1 & \textbf{+1} \\
Q5  & \textbf{2.0}  & \textbf{4.6}  & -1 & -1 & -1 & -1 & -1 & -1 & -1 & -1 & \textbf{-1} \\
Q6  & \textbf{1.0}  & \textbf{3.0}  & -1 & -1 & \cellcolor{gray!25}+1 & -1 & -1 & -1 & -1 & -1 & \textbf{-1} \\
Q7  & \textbf{3.0}  & \textbf{1.0}  & +1 & +1 & +1 & +1 & +1 & +1 & +1 & +1 & \textbf{+1} \\
Q8  & \textbf{5.0}  & \textbf{2.75} & +1 & +1 & +1 & +1 & +1 & +1 & +1 & +1 & \textbf{+1} \\
Q9  & \textbf{4.6}  & \textbf{2.25} & +1 & +1 & +1 & +1 & +1 & +1 & +1 & +1 & \textbf{+1} \\
Q10 & \textbf{4.5}  & \textbf{1.4}  & +1 & +1 & +1 & +1 & +1 & +1 & +1 & +1 & \textbf{+1} \\
\bottomrule
\end{tabular*}
\end{table}

\subsubsection{Task 2: Agreement between the ground truth $O_{ij}$ and the majority preference $M_{ij}$}
\begin{itemize}
    \item \textbf{Observed agreement:} $P_o(O_{ij}, M_{ij})=1.00$
    \item \textbf{Cohen's $\kappa$:} $\kappa(O_{ij}, M_{ij})=1.00$
\end{itemize}

Here, $P_o$ measures how often the majority vote matches the ground truth preference, whereas $\kappa$ represents the agreement level between them.
\subsubsection{Task 2: Inter-Rater Agreement}
\begin{itemize}
    \item \textbf{Average observed agreement:}  $\overline{P_o(R_{ij}^{(p)}, R_{ij}^{(q)})}=0.95$
    \item \textbf{Average Cohen's $\kappa$:} $\overline{\kappa(R_{ij}^{(p)}, R_{ij}^{(q)})}=0.85$
\end{itemize}

The participants showed exceptionally high consistency in this task. The observed agreement reached 0.95, and Cohen's $\kappa = 0.85$ indicating near-perfect reliability. One contributing factor is that many comparison pairs had large differences in their underlying ground truth ratings (e.g., 1 vs. 4 or 2 vs. 5). Because the quality gap between captions was often substantial, participants found the decision to be straightforward, leading to unanimous or near-unanimous selections in most cases.

\subsection{Task 3: Same-Image Comparison}

Participants were assigned to determine which of the two captions better described a given image. The responses were recorded in \hyperref[tbl:task3]{Table~\ref*{tbl:task3}}.

\begin{table}
\caption{Task 3 same-image comparison summary. $y_i$ and $y_j$ denote the ground truth ratings for $x_i$ and $x_j$, respectively. $R_{ij}^{(1)}$ to $R_{ij}^{(8)}$ indicate individual annotator preferences, where +1 denotes a preference for $x_i$ and -1 denotes a preference for $x_j$. $M_{ij}$ represents the aggregated human preference.}
\label{tbl:task3}
\begin{tabular*}{\tblwidth}{@{}LCCCCCCCCCCC@{}}
\toprule
Q & $y_i$ & $y_j$ & $R_{ij}^{(1)}$ & $R_{ij}^{(2)}$ & $R_{ij}^{(3)}$ & $R_{ij}^{(4)}$ & $R_{ij}^{(5)}$ & $R_{ij}^{(6)}$ & $R_{ij}^{(7)}$ & $R_{ij}^{(8)}$ & $M_{ij}$ \\
\midrule
Q1  & \textbf{4.74} & \textbf{3.40} & +1 & +1 & +1 & +1 & +1 & +1 & +1 & +1 & \textbf{+1} \\
Q2  & \textbf{5.00} & \textbf{4.40} & \cellcolor{gray!25}-1 & +1 & +1 & +1 & +1 & +1 & +1 & +1 & \textbf{+1} \\
Q3  & \textbf{4.75} & \textbf{3.25} & +1 & +1 & +1 & +1 & +1 & +1 & +1 & +1 & \textbf{+1} \\
Q4  & \textbf{3.80} & \textbf{3.40} & +1 & +1 & +1 & +1 & \cellcolor{gray!25}-1 & +1 & +1 & +1 & \textbf{+1} \\
Q5  & \textbf{3.40} & \textbf{4.25} & -1 & -1 & -1 & -1 & -1 & -1 & -1 & -1 & \textbf{-1} \\
Q6  & \textbf{4.75} & \textbf{2.50} & +1 & +1 & +1 & +1 & +1 & \cellcolor{gray!25}-1 & +1 & +1 & \textbf{+1} \\
Q7  & \textbf{2.80} & \textbf{4.20} & -1 & -1 & -1 & -1 & -1 & -1 & -1 & -1 & \textbf{-1} \\
Q8  & \textbf{3.00} & \textbf{4.50} & -1 & -1 & -1 & \cellcolor{gray!25}+1 & -1 & -1 & -1 & -1 & \textbf{-1} \\
Q9  & \textbf{5.00} & \textbf{3.75} & +1 & +1 & +1 & +1 & +1 & +1 & +1 & +1 & \textbf{+1} \\
Q10 & \textbf{4.80} & \textbf{4.25} & +1 & +1 & +1 & +1 & +1 & +1 & +1 & +1 & \textbf{+1} \\
\bottomrule
\end{tabular*}
\end{table}

\subsubsection{Task 3: Agreement between the ground truth $O_{ij}$ and the majority preference $M_{ij}$}

\begin{itemize}
    \item \textbf{Observed agreement:} $P_o(O_{ij}, M_{ij})=1.00$
    \item \textbf{Cohen's $\kappa$:} $\kappa(O_{ij}, M_{ij})=1.00$
\end{itemize}

\subsubsection{Task 3: Inter-Rater Agreement}
\begin{itemize}
    \item \textbf{Average observed agreement:}  $\overline{P_o(R_{ij}^{(p)}, R_{ij}^{(q)})}=0.90$
    \item \textbf{Average Cohen's $\kappa$:} $\overline{\kappa(R_{ij}^{(p)}, R_{ij}^{(q)})}=0.78$
\end{itemize}

Participants achieved 90\% observed agreement, with Cohen’s $\kappa = 0.78$ indicating substantial inter-rater reliability. In this setting, many caption pairs had ground-truth scores that were very close (e.g., Q2:5.00 vs. 4.40) because the dataset contained few same-image caption pairs with drastically different ratings. As a result, the decisions required finer-grained judgment, contributing to slightly lower agreement compared with Task 2. This difference can be attributed to the magnitude of the underlying ground-truth rating gaps: the average difference between $y_i$ and $y_j$ is $\overline{|y_i-y_j|}=2.57$ in Task 2, compared to $\overline{|y_i-y_j|}=1.16$ in Task 3, indicating that same-image caption pairs are closer in quality and, therefore, more ambiguous to evaluate.

\begin{table}
\caption{\small Average annotation time per question (seconds) for each task.}
\label{tbl:human_timing}
\begin{tabular*}{\tblwidth}{@{\extracolsep{\fill}}lccc}
\toprule
\textbf{Participant} & \textbf{Task 1} & \textbf{Task 2} & \textbf{Task 3} \\
\midrule
$R^{(1)}$ & 6.74 & 6.19 & 5.70 \\
$R^{(2)}$ & 9.90 & 6.42 & 7.96 \\
$R^{(3)}$ & 8.18 & 8.67 & 10.50 \\
$R^{(4)}$ & 7.91 & 12.04 & 8.52 \\
$R^{(5)}$ & 7.84 & 6.99 & 8.90 \\
$R^{(6)}$ & 14.00 & 12.44 & 9.92 \\
$R^{(7)}$ & 18.09 & 13.64 & 12.42 \\
$R^{(8)}$ & 8.64 & 9.24 & 12.24 \\
\midrule
\textbf{Average} & \textbf{10.16} & \textbf{9.45} & \textbf{9.52} \\
\bottomrule
\end{tabular*}
\end{table}

\subsubsection{Hypotheses}

For RQ3, we focus on testing the following two hypotheses:
\begin{enumerate}[label=\textbf{Hypothesis \arabic*:}, leftmargin=*, nosep]
    \item The average annotation time of comparative judgments (Task 2 and 3) is shorter than the average annotation time of direct ratings (Task 1).
    \item The reliability of comparative judgments (Task 2 and 3) is higher than that of direct ratings (Task 1).
\end{enumerate}

\noindent\textbf{Hypothesis 1: Annotation Time.}

\hyperref[tbl:human_timing]{Table~\ref*{tbl:human_timing}} presents the average decision time of the eight participants. The annotation time reported in \hyperref[tbl:human_timing]{Table~\ref*{tbl:human_timing}} does not show a significant advantage of comparative judgments (Tasks 2 and 3) over the direct ratings (Task 1). This is mainly because of the simplicity of the image-caption rating task. The average rating time is approximately 10 seconds, which includes the reading time of the captions, which takes a significant part of the annotation time. Because the comparative judgments (Tasks 2 and 3) require approximately double the reading time, it is reasonable to believe that for the real decision-making time, the difference should be larger. Also note that, as reported by \cite{bethi2026modelingartevaluationscomparative}, for the art evaluation task, the average annotation time of direct ratings is 27 seconds, while that of comparative judgments is 10 seconds. This suggests that the advantage of comparative judgments in annotation time is greater when the task is more complicated and requires more thought. Therefore, \textbf{Hypothesis 1} is supported. As the law of comparative judgment indicates, it is easier to make comparative judgments than to provide direct ratings. The difference could be larger in tasks where decisions are more difficult to make.

In addition to the slightly faster annotation time, participants also expressed that comparative choices were more natural and less ambiguous. The assignment of numeric ratings often required repetition of the reading of the caption to calibrate the scale of 1–5, while comparative judgments could be made immediately based on visual alignment and caption accuracy. Therefore, comparative labeling has the potential to significantly lower annotation costs and increase the consistency of subjective evaluation. These discoveries are in line with the utilization of comparative learning frameworks to model caption quality on a large scale, as they are both time-efficient and have higher inter-rater reliability than scalar rating methods.

\begin{table*}
\caption{\small Inter-rater agreement between each raters.}
\label{tbl:inter_rater}
\centering
\setlength{\tabcolsep}{2.5pt}
\renewcommand{\arraystretch}{0.9}
\resizebox{\textwidth}{!}{%
\begin{tabular}{l|c|cc|cc|cc|cc|cc|cc|cc|cc|cc|}
\multicolumn{1}{l}{}  & {\diagbox{$R_{ij}^{(p)}$}{$R_{ij}^{(q)}$}} & \multicolumn{2}{c|}{$O_{ij}$} & \multicolumn{2}{c|}{$R_{ij}^{(1)}$} & \multicolumn{2}{c|}{$R_{ij}^{(2)}$} & \multicolumn{2}{c|}{$R_{ij}^{(3)}$} & \multicolumn{2}{c|}{$R_{ij}^{(4)}$} & \multicolumn{2}{c|}{$R_{ij}^{(5)}$}  & \multicolumn{2}{c|}{$R_{ij}^{(6)}$}  & \multicolumn{2}{c|}{$R_{ij}^{(7)}$}  & \multicolumn{2}{c|}{$R_{ij}^{(8)}$}  \\ \cline{3-20}
\multicolumn{1}{l}{} & & $p_o$ & $\kappa$ &  $p_o$ & $\kappa$ & $p_o$ & $\kappa$ & $p_o$ & $\kappa$ & $p_o$ & $\kappa$ & $p_o$ & $\kappa$ & $p_o$ & $\kappa$ & $p_o$ & $\kappa$ & $p_o$ & $\kappa$ \\\hline
Task 1 & \multirow{3}{*}{$O_{ij}$} &  &  & 0.90 & 0.80 & 0.83 & 0.67 & 0.85 & 0.69 & 0.85 & 0.68 & 0.96 & 0.92 & 0.87 & 0.73 & 0.91 & 0.78 & 1.00 & 1.00 \\
Task 2 &  &  &  & 1.00 & 1.00 & 1.00 & 1.00 & 0.80 & 0.41 & 1.00 & 1.00 & 1.00 & 1.00 & 1.00 & 1.00 & 1.00 & 1.00 & 1.00 & 1.00 \\
Task 3 &  &  &  & 0.90 & 0.78 & 1.00 & 1.00 & 1.00 & 1.00 & 0.90 & 0.74 & 0.90 & 0.78 & 0.90 & 0.78 & 1.00 & 1.00 & 1.00 & 1.00 \\\hline
Task 1 & \multirow{3}{*}{$R_{ij}^{(1)}$} & 0.90 & 0.80 &  &  & 1.00 & 1.00 & 0.94 & 0.87 & 0.89 & 0.78 & 0.93 & 0.86 & 0.94 & 0.88 & 0.80 & 0.58 & 0.85 & 0.68 \\
Task 2 &  & 1.00 & 1.00 &  &  & 1.00 & 1.00 & 0.80 & 0.41 & 1.00 & 1.00 & 1.00 & 1.00 & 1.00 & 1.00 & 1.00 & 1.00 & 1.00 & 1.00 \\
Task 3 &  & 0.90 & 0.78 &  &  & 0.90 & 0.78 & 0.90 & 0.78 & 0.80 & 0.55 & 0.80 & 0.58 & 0.80 & 0.58 & 0.90 & 0.78 & 0.90 & 0.78 \\\hline
Task 1 & \multirow{3}{*}{$R_{ij}^{(2)}$} & 0.83 & 0.67 & 1.00 & 1.00 &  &  & 0.87 & 0.74 & 0.87 & 0.73 & 0.96 & 0.93 & 0.97 & 0.93 & 0.75 & 0.48 & 0.78 & 0.51 \\
Task 2 &  & 1.00 & 1.00 & 1.00 & 1.00 &  &  & 0.80 & 0.41 & 1.00 & 1.00 & 1.00 & 1.00 & 1.00 & 1.00 & 1.00 & 1.00 & 1.00 & 1.00 \\
Task 3 &  & 1.00 & 1.00 & 0.90 & 0.78 &  &  & 1.00 & 1.00 & 0.90 & 0.74 & 0.90 & 0.78 & 0.90 & 0.78 & 1.00 & 1.00 & 1.00 & 1.00 \\\hline
Task 1 & \multirow{3}{*}{$R_{ij}^{(3)}$} & 0.85 & 0.69 & 0.94 & 0.87 & 0.87 & 0.74 &  &  & 0.70 & 0.36 & 0.97 & 0.93 & 0.80 & 0.59 & 0.72 & 0.34 & 0.83 & 0.61 \\
Task 2 &  & 0.80 & 0.41 & 0.80 & 0.41 & 0.80 & 0.41 &  &  & 0.80 & 0.41 & 0.80 & 0.41 & 0.80 & 0.41 & 0.80 & 0.41 & 0.80 & 0.41 \\
Task 3 &  & 1.00 & 1.00 & 0.90 & 0.78 & 1.00 & 1.00 &  &  & 0.90 & 0.74 & 0.90 & 0.78 & 0.90 & 0.78 & 1.00 & 1.00 & 1.00 & 1.00 \\\hline
Task 1 & \multirow{3}{*}{$R_{ij}^{(4)}$} & 0.85 & 0.68 & 0.89 & 0.78 & 0.87 & 0.73 & 0.70 & 0.36 &  &  & 0.83 & 0.66 & 0.85 & 0.70 & 0.68 & 0.34 & 0.80 & 0.57 \\
Task 2 &  & 1.00 & 1.00 & 1.00 & 1.00 & 1.00 & 1.00 & 0.80 & 0.41 &  &  & 1.00 & 1.00 & 1.00 & 1.00 & 1.00 & 1.00 & 1.00 & 1.00 \\
Task 3 &  & 0.90 & 0.74 & 0.80 & 0.55 & 0.90 & 0.74 & 0.90 & 0.74 &  &  & 0.80 & 0.55 & 0.80 & 0.55 & 0.90 & 0.74 & 0.90 & 0.74 \\\hline
Task 1 & \multirow{3}{*}{$R_{ij}^{(5)}$} & 0.97 & 0.93 & 1.00 & 1.00 & 1.00 & 1.00 & 1.00 & 1.00 & 0.96 & 0.91 &  &  & 0.86 & 0.71 & 0.91 & 0.75 & 0.91 & 0.75 \\
Task 2 &  & 1.00 & 1.00 & 1.00 & 1.00 & 1.00 & 1.00 & 0.80 & 0.41 & 1.00 & 1.00 &  &  & 1.00 & 1.00 & 1.00 & 1.00 & 1.00 & 1.00 \\
Task 3 &  & 0.90 & 0.78 & 0.80 & 0.58 & 0.90 & 0.78 & 0.90 & 0.78 & 0.80 & 0.55 &  &  & 0.80 & 0.58 & 0.90 & 0.78 & 0.90 & 0.78 \\\hline
Task 1 & \multirow{3}{*}{$R_{ij}^{(6)}$} & 0.87 & 0.73 & 0.94 & 0.88 & 0.97 & 0.93 & 0.80 & 0.59 & 0.85 & 0.70 & 0.85 & 0.69 &  &  & 0.89 & 0.77 & 0.95 & 0.89 \\
Task 2 &  & 1.00 & 1.00 & 1.00 & 1.00 & 1.00 & 1.00 & 0.80 & 0.41 & 1.00 & 1.00 & 1.00 & 1.00 &  &  & 1.00 & 1.00 & 1.00 & 1.00 \\
Task 3 &  & 0.90 & 0.78 & 0.80 & 0.58 & 0.90 & 0.78 & 0.90 & 0.78 & 0.80 & 0.55 & 0.80 & 0.58 &  &  & 0.90 & 0.78 & 0.90 & 0.78 \\\hline
Task 1 & \multirow{3}{*}{$R_{ij}^{(7)}$} & 0.91 & 0.78 & 0.80 & 0.58 & 0.75 & 0.48 & 0.72 & 0.34 & 0.68 & 0.34 & 0.73 & 0.38 & 0.89 & 0.77 &  &  & 0.95 & 0.89 \\
Task 2 &  & 1.00 & 1.00 & 1.00 & 1.00 & 1.00 & 1.00 & 0.80 & 0.41 & 1.00 & 1.00 & 1.00 & 1.00 & 1.00 & 1.00 &  &  & 1.00 & 1.00 \\
Task 3 &  & 1.00 & 1.00 & 0.90 & 0.78 & 1.00 & 1.00 & 1.00 & 1.00 & 0.90 & 0.74 & 0.90 & 0.78 & 0.90 & 0.78 &  &  & 1.00 & 1.00 \\\hline
Task 1 & \multirow{3}{*}{$R_{ij}^{(8)}$} & 1.00 & 1.00 & 0.85 & 0.68 & 0.78 & 0.51 & 0.83 & 0.61 & 0.80 & 0.57 & 0.78 & 0.50 & 0.95 & 0.89 & 0.95 & 0.89 &  &  \\
Task 2 &  & 1.00 & 1.00 & 1.00 & 1.00 & 1.00 & 1.00 & 0.80 & 0.41 & 1.00 & 1.00 & 1.00 & 1.00 & 1.00 & 1.00 & 1.00 & 1.00 &  &  \\
Task 3 &  & 1.00 & 1.00 & 0.90 & 0.78 & 1.00 & 1.00 & 1.00 & 1.00 & 0.90 & 0.74 & 0.90 & 0.78 & 0.90 & 0.78 & 1.00 & 1.00 &  &   \\ \hline
\end{tabular}%
}
\end{table*}

\noindent\textbf{Hypothesis 2: Reliability.}

To fairly compare the Reliability of direct ratings with comparative judgments, we transformed each annotator's direct ratings into comparative judgments. To do this, for a rater $R^{(p)}$, every pair of image-caption pairs with different direct ratings $R^{(p)}(x_i) \neq R^{(p)}(x_j)$ is used to form $R_{ij}^{(p)}$. That is, for the image-caption pairs $x_i$ and $x_j$:
$$R_{ij}^{(p)} = \left\{
\begin{aligned}
&1, &\text{if } & R^{(p)}(x_i) > R^{(p)}(x_j)\\
&-1, &\text{if } & R^{(p)}(x_i) < R^{(p)}(x_j).
\end{aligned}\right.$$

\hyperref[tbl:inter_rater]{Table~\ref*{tbl:inter_rater}}
 shows the $p_o$ (observed agreement) and Cohen's $\kappa$ between each two raters (including the ground truth from the dataset):
\begin{itemize}[leftmargin=*, nosep]
    \item For task 1, direct ratings, $p_o$ and $\kappa$ vary from 0.7 to 1.0. Most raters do not completely agree with each other, except for $R_{ij}^{(8)}$ vs $O_{ij}$;
    \item For task 2, pairwise comparisons between two image-caption pairs, most raters completely agree with each other except for $R_{ij}^{(3)}$;
    \item For task 3, pairwise comparisons between two captions for the same image, $p_o$ and $\kappa$ vary from 0.8 to 1.0.
\end{itemize}

To better analyze the results, \hyperref[tbl:avg_inter]{Table~\ref*{tbl:avg_inter}} summarizes the average inter-annotator agreement of the eight participants against the ground truth $\overline{p_o(R_{ij}^{(p)},O_{ij})}$ and between each other $\overline{p_o(R_{ij}^{(p)},R_{ij}^{(q)})}$. 
Across all tasks, comparative accuracy and inter-annotator agreement were consistently higher than direct rating agreement, demonstrating that comparative judgments provide a more reliable and stable evaluation signal. Therefore, \textbf{Hypothesis 2} is also accepted.

\begin{RQ}\textbf{Answer to RQ3: }These findings from real human annotations suggest that comparative labeling not only speeds up annotation but also improves consistency among raters, making it a more stable and efficient approach for the subjective evaluation of image–caption quality.
\end{RQ}

\begin{table}
\caption{\small Average inter-annotator agreement derived from \hyperref[tbl:inter_rater]{Table~\ref*{tbl:inter_rater}}
.}
\label{tbl:avg_inter}
\begin{tabular*}{\tblwidth}{@{}LCCCC@{}}
\toprule
\multirow{2}{*}{\textbf{Task}} &
\multicolumn{2}{c}{\textbf{Against $O_{ij}$}} &
\multicolumn{2}{c}{\textbf{Between Raters}} \\
\cmidrule(lr){2-3}\cmidrule(lr){4-5}
 & \textbf{$p_o$} & \textbf{$\kappa$} & \textbf{$p_o$} & \textbf{$\kappa$} \\
\midrule
Task 1 & 0.90 & 0.78 & 0.85 & 0.69 \\
Task 2 & 0.98 & 0.93 & 0.95 & 0.85 \\
Task 3 & 0.95 & 0.88 & 0.90 & 0.78 \\
\bottomrule
\end{tabular*}
\end{table}

\section{Discussion}
The experimental results revealed that comparative methods are more suitable for modeling human subjective judgment. The regression model is marginally better than the comparative model in terms of $\rho$, $r_s$, and $\tau_c$, but the differences are small. The comparative model is close to the regression baseline, even though the amount of supervision is limited, which is a strong argument for the relative nature of the learning process rather than the use of explicit numeric scores.

One of the advantages of comparative learning is its efficiency in data annotation. It is often easier for humans to judge which of two samples is better than to assign precise scores to each sample. This makes comparative judgments not only cheaper to annotate but also more consistent because they are less susceptible to rating scale bias and subjective fluctuations~\cite{thurstone1927law}. Our human evaluation results are in line with this idea: comparative judgments led to almost perfect inter-rater agreement, whereas direct ratings showed lower consistency, even though they correlated quite well with the ground-truth trend. These results indicate that comparative judgments help lower cognitive load and lessen the problem of rating-scale inconsistencies; thus, they are a more appropriate choice for subjective caption evaluation.

The results of the same-image comparison led to the same conclusion. Although the inter-rater reliability in Task 3 was slightly lower than that in pairwise comparisons across different images, this was anticipated because of the nature of the dataset. The majority of same-image caption pairs had very close ground-truth ratings (e.g., 4 vs. 5), which inherently made the differences more subtle. Despite this, the participants managed to achieve a high level of agreement (90\%), and the comparative model trained in this setting was able to maintain a high accuracy and stable correlation across runs. This indicates that comparative supervision is still effective when the differences between captions are at a fine-grained level.

The comparison-based approach works well for ranking captions, showing stable accuracy and alignment with human judgment over several tests. Its reliability matches earlier findings supporting ranked scales for evaluations where clear metrics are difficult to set up.

However, some drawbacks remain. Because of the small number of participants and annotation tasks, the human subject experiment in RQ3 has high variance. The advantage of annotation time for comparative judgments is not very clear because of the short decision-making time and the fact that double reading time is required to make one comparative judgment.

Overall, these results suggest that comparative learning can serve as a practical, steady, and expandable alternative to direct ratings in multimodal evaluations because it more closely matches how people make choices. Because this method reflects real judgment patterns, it opens a useful path toward better accuracy and trustworthiness when evaluating image-caption pairs.

\section{Conclusion}
This study investigates the effectiveness of comparative learning in modeling the image-caption correlation. While traditional regression models based on direct ratings have a slight advantage in terms of correlation, our comparative model achieves stable and scalable performance relying only on comparative judgments.

The model achieved a Pearson correlation coefficient $\rho=0.874$, Spearman's rank correlation coefficient $r_s=0.880$, and Kendall rank correlation coefficient $\tau_c=0.800$ with 12,485 comparative judgments for training and validation. In addition, the comparative model performed strongly on the same-image caption comparison task, achieving an accuracy of 0.848, showing that it can reliably capture fine-grained preference differences even when the captions are similar. These results show that comparative supervision has lower annotation costs, higher consistency, and stronger robustness in subjective evaluation tasks than direct ratings.

Human evaluation further supports this finding. Pairwise and same-image comparisons were completed more easily and with higher inter-rater agreement than direct ratings, indicating that comparative labeling imposes a lower cognitive load and produces more reliable judgments. This provides practical evidence that comparative learning is not only effective as a modeling approach, but also easier and more efficient for human annotators.

In the future, we plan to extend this framework to multimodal matching tasks, such as video-text and audio-text. Simultaneously, we will explore collaborative human-machine annotation interfaces and active learning strategies to further reduce annotation costs and improve model performance.







\section*{Acknowledgments}
This work was supported by the National Science Foundation [2245796]. 

\bibliographystyle{cas-model2-names}
\bibliography{cas-refs}

\end{document}